\documentclass[conference]{IEEEtran}
\IEEEoverridecommandlockouts
\usepackage{cite}
\usepackage{amsmath,amssymb,amsfonts}
\usepackage{algorithmic}
\usepackage{graphicx}
\usepackage{textcomp}
\usepackage{xcolor}
\usepackage{float}
\usepackage{tabularx}
\usepackage{booktabs}
\usepackage{subcaption}
\usepackage{url}
\def\BibTeX{{\rm B\kern-.05em{\sc i\kern-.025em b}\kern-.08em
    T\kern-.1667em\lower.7ex\hbox{E}\kern-.125emX}}
\begin{document}

\title{Hands-on Evaluation of Visual Transformers for Object Recognition and Detection\\
{\footnotesize \textsuperscript{}}
\thanks{}
}

\author{\IEEEauthorblockN{Anonymous Authors}}

\author{\IEEEauthorblockN{Dimitrios N. Vlachogiannis}
\IEEEauthorblockA{\textit{Dept. of Computer Engineering and Informatics} \\
\textit{ University of Patras}\\
Patras, Greece \\
st1067371@ceid.upatras.gr}
\and
\IEEEauthorblockN{Dimitrios A. Koutsomitropoulos}
\IEEEauthorblockA{\textit{Dept. of Computer Engineering and Informatics} \\
\textit{University of Patras}\\
 Patras, Greece \\
 koutsomi@ceid.upatras.gr}
\and
}

\maketitle

\begin{abstract}
Convolutional Neural Networks (CNNs) for computer vision sometimes struggle with understanding images in a global context, as they mainly focus on local patterns. On the other hand,  Vision Transformers (ViTs), inspired by models originally created for language processing, use self-attention mechanisms, which allow them to understand relationships across the entire image.
In this paper, we compare different types of ViTs (pure, hierarchical, and hybrid) against traditional CNN models across various tasks, including object recognition, detection, and medical image classification. We conduct thorough tests on standard datasets like ImageNet for image classification and COCO for object detection. Additionally, we apply these models to medical imaging using the ChestX-ray14 dataset. We find that hybrid and hierarchical transformers, especially Swin and CvT, offer a strong balance between accuracy and computational resources. Furthermore, by experimenting with data augmentation techniques on medical images, we discover significant performance improvements, particularly with the Swin Transformer model. Overall, our results indicate that Vision Transformers are competitive and, in many cases, outperform traditional CNNs, especially in scenarios requiring the understanding of global visual contexts like medical imaging.
\end{abstract}


\section{Introduction}
The rapid development of deep learning and artificial intelligence over the last decade has transformed computer vision, enabling machines to interpret and understand images and videos with greater accuracy than ever before. Convolutional Neural Networks (CNNs) have been the default approach to visual recognition tasks for over a decade, doing both image classification and object detection due to their ability to learn local spatial hierarchies through convolutional operations. Architectures such as AlexNet \cite{krizhevsky2012imagenet}, VGGNet \cite{simonyan2015very}, ResNet \cite{he2016deep} for object recognition and Faster R-CNN \cite{ren2015faster}, Yolo \cite{redmon2016you} for object detection have established CNNs as the dominant architecture in computer vision. However, CNNs often face limitations related to capturing global context effectively, which is crucial for a more in-depth insight into visual scenes and robust performance, especially in challenging conditions \cite{paul2023are, benz2021adversarial}.

The Transformer model, as presented by Vaswani et al. \cite{vaswani2017attention}, has revolutionized the domain of natural language processing (NLP) by effectively capturing long-range dependencies through the self-attention mechanism, thus enabling models like BERT \cite{devlin2019bert} and GPT \cite{radford2018improving} to achieve state-of-the-art results. Inspired by these  successes, the Vision Transformer (ViT) \cite{dosovitskiy2021image} was devised, adapting Transformer models to vision tasks by framing images as sequences of image patches processed via self-attention. Unlike CNNs, ViTs inherently capture global image features and significantly improve their capacity to identify spatial relationships present across the entire image. 

Furthermore, the robustness and application of Vision Transformers in specialized domains, notably medical imaging, have become important topics. Transformers have demonstrated enhanced resilience against adversarial perturbations compared to CNNs \cite{bhojanapalli2021adversarial,paul2023are, dey2023exploring}, attributed to their global context capabilities. Additionally, in medical imaging scenarios, ViTs have showcased superior performance, reduced sensitivity to hidden stratification, and improved generalization across diverse datasets \cite{wu2021vision, gheflati2022vision, chu2022visual}.

Motivated by these latest developments, this paper presents a comprehensive hands-on evaluation of various Vision Transformer architectures for object recognition, detection, and medical image analysis tasks.The main contributions of this work are the following:

\begin{itemize}
    \item Comprehensive Comparative Evaluation: We present an extensive experimental evaluation and comparative analysis of Transformer-based architecture (pure, hierarchical, and hybrid ViT models) against established CNN benchmarks on both image classification (ImageNet-1K) and object detection (COCO) datasets.
    \item We extend our evaluation to medical image classification, utilizing the ChestX-ray14 dataset. We demonstrate the effectiveness of Vision Transformers, particularly that of hybrid and hierarchical ViTs. 
    \item We investigate specific data augmentation techniques~(CutMix, MixUp, Random Augmentations) and observe their impact on the hierarchical model (Swin). Notably, these techniques have not been previously applied to the pure Swin model on the ChestX-ray14 dataset.
    \item Reproducible Experimental Framework: All experiments are performed using publicly accessible tools and platforms (HuggingFace Transformers and Trainer), with publicly available code \cite{tsomaros_vit} and model checkpoints.
\end{itemize}

The rest of this paper is organized as follows: In the next Section II we review related work on ViTs and present a comparative evaluation of the different models based on the literature, differentiating between pure Transformer architectures and hybrid approaches. In Section III we discuss our evaluation methodology and present evaluation results for the two tasks of object recognition and detection accordingly, comparing performance between the different ViT architectures as well as traditional CNNs. In Section IV we focus on the medical imaging problem domain and evaluate the most promising architecture representatives found in the previous section on object detection in the ChestXray14 dataset. The potential of data augmentation techniques on the best model (Swin-base) is also investigated and reported. Finally, Section V summarizes our conclusions and future work.

\section{Background and Related Work}
\subsection{ViT for Object Recognition}
ViT \cite{dosovitskiy2021image} has demonstrated excellent results in various image recognition benchmarks (ImageNet \cite{deng2009imagenet}, CIFAR-100 \cite{krizhevsky2009learning}) and competitive results compared with state-of-the-art CNN models. By taking advantage of the self-attention mechanism to incorporate information from the entire image from the earliest stages, it achieves a better understanding of global correlations compared to CNNs that rely mainly on local filters \cite{dosovitskiy2021image, touvron2021training, paul2023are}. In addition, ViT offers increased flexibility in handling high-resolution images through patch processing \cite{dosovitskiy2021image}, while exhibiting greater robustness to noise and corruption thanks to global feature integration \cite{bhojanapalli2021adversarial, paul2023are,  dey2023exploring}. Despite its benefits, ViT encounters some drawbacks. This includes the significant computational complexity linked to its self-attention mechanism that demands substantial computational resources \cite{dosovitskiy2021image}. Moreover, because transformers lack inherent inductive biases, ViT generally requires extensive datasets to learn efficiently \cite{touvron2021training}. Finally, ViT frequently shows deficiencies in capturing local features, impacting its effectiveness in tasks like small object detection \cite{carion2020end}.

To limit the problems that the original ViT had, certain variations have been proposed which aim to enhance the locality and the self-attention mechanism. PVT \cite{wang2021pyramid} is a hierarchical version of ViT, implementing multiple resolution scales, allowing for efficient modeling of details at different levels. This approach makes it particularly effective in classifying high-resolution images, improving accuracy with lower computational cost. Swin Transformer \cite{liu2021swin} introduced the shifted windows mechanism, reducing the self-attention cost from quadratic to linear. With this approach, it maintained high classification accuracy, making it one of the most efficient Transformer-based architectures for visual tasks. CvT \cite{wu2021cvt} is a hybrid model that incorporates convolutional operations in the early stages of the Transformer to exploit the powerful local modeling of CNNs. This hybrid approach achieves high performance in image classification while simultaneously leveraging the advantages of CNNs and Transformers. LeViT \cite{graham2021levit} combines the advantages of CNNs and Transformers in a lightweight architecture, designed to provide efficient classification with a reduced number of parameters and computational complexity. The incorporation of convolutions in the early stages makes it ideal for applications requiring high speed and low resource usage. Table \ref{tab:paper_model_comparison} shows a comparison between all the presented works according to their size, computational requirements and top-1 accuracy on ImageNet.

\begin{table}[htbp]
\centering
\caption{ImageNet result comparison of different vision transformer architectures}
\renewcommand{\arraystretch}{1.1} 
\setlength{\tabcolsep}{8pt}
\begin{tabular}{lccc}
\hline
\textbf{Model} & \textbf{\#Params (M)} & \textbf{FLOPs (G)} & \textbf{Top-1(\%)} \\ \hline
\multicolumn{4}{c}{\textbf{Pure Transformer}} \\ \hline
ViT-B\cite{dosovitskiy2021image}         & 86.6 & 17.6  & 83.9  \\
ViT-L\cite{dosovitskiy2021image}        & 307 & 61.6  & 85.1   \\
PvT-Medium\cite{wang2021pyramid}        & 44.2 & 6.7 & 81.2   \\
PvT-Large\cite{wang2021pyramid}  & 61.4   & 9.8 & 81.7  \\
Swin-B\cite{liu2021swin}  & 88   & 15.4 & 83.3  \\
Swin-L\cite{liu2021swin}  & 197   & 104   & 86.4  \\ \hline
\multicolumn{4}{c}{\textbf{Hybrid}} \\ \hline
ViT-Hybrid-B\cite{dosovitskiy2021image}           & 99     & 49.6   & 85.5  \\
CvT-21\cite{wu2021cvt}                & 32     & 7.1  & 82.5  \\
CvT-21-384\cite{wu2021cvt}                & 32   & 24.9  & 83.3    \\
LeViT-256\cite{graham2021levit}                 & 18.9   & 1.12 & 81.6   \\
LeViT-384\cite{graham2021levit}                 & 39.1   & 2.35 & 82.6   \\ \hline
\end{tabular}
\label{tab:paper_model_comparison}
\end{table}

\subsection{ViT for Object Detection}
Traditional object detectors are mainly based on CNNs, but after the exceptional performance shown by transformers in the field of classification, they began to be introduced into object detection. They can be categorized into two groups:

\subsubsection{Detection Transformers with CNN Backbone} DETR \cite{carion2020end} was the first fully Transformer-based model for object detection, introducing an end-to-end detection mechanism without the need for anchors or region proposal networks. Although simple in design, it exhibited relatively slow convergence and poor performance on small objects. Deformable DETR \cite{zhu2021deformable} is an improvement on DETR by introducing deformable attention that focuses only on a small number of critical points. This drastically reduces training time and improves performance on small objects. Swin Transformer \cite{liu2021swin} is often used as a backbone in object detection systems such as Cascade Mask R-CNN, offering hierarchical features with window-based self-attention. Its use allows state-of-the-art performance on the COCO dataset, with very good scaling in depth and resolution.

\subsubsection{Detection with Pure Transformers} YOLOS \cite{fang2021yolos} is an application of ViT directly to object detection, treating the input as a sequence of patches, without any special adaptation for localization. Although simple, it lacks accuracy compared to more specialized architectures but shows that pure Vision Transformers have capabilities beyond classification. Table \ref{tab:paper_object_detection} shows a comparison between the above models according to their size, computational requirements and average precision (AP) for object detection on COCO 2017 validation set.

\begin{table}[htbp]
\centering
\caption{Performance comparison of transformer-based object detectors on the COCO 2017 validation set.}
\renewcommand{\arraystretch}{1.7} 
\resizebox{\columnwidth}{!}{%
\begin{tabular}{lrrrrr}
\toprule
\textbf{Model} & \textbf{AP} & \textbf{AP$_{50}$} & \textbf{AP$_{75}$} & \textbf{\#Params (M)} & \textbf{FLOPs (G)} \\
\midrule
\multicolumn{6}{c}{\textbf{Pure Transformer}} \\
\midrule
YOLOS-Tiny~\cite{fang2021yolos} & 0.300 &   --   &   --  &  6.5 &   21 \\
YOLOS-Base~\cite{fang2021yolos} & 0.420 & 0.622 & 0.445 &  127.8 &  537 \\
\midrule
\multicolumn{6}{c}{\textbf{CNN Backbone (Hybrid)}} \\
\midrule
DETR-R50~\cite{carion2020end}           & 0.420 & 0.624 & 0.442 &  41.3 &   86 \\
DETR-R101~\cite{carion2020end}          & 0.433 & 0.631 & 0.459 &  41.0 &  187 \\
Cond.-DETR-R50~\cite{meng2021conditional} & 0.430 & 0.640 & 0.457 &  44.0 &   90 \\
Def.-DETR~\cite{zhu2021deformable}      & 0.462 & 0.652 & 0.500 &  40.0 &  173 \\
\bottomrule
\end{tabular}%
}
\label{tab:paper_object_detection}
\end{table}

\subsection{Special Features of ViTs}
In addition to ViT performance on basic problems (classification, object detection), there is particular interest in studying their robustness against perturbation, as well as their utilization in medical imaging applications. These two features occur from the ability of ViTs to capture global dependencies through the self-attention mechanism, which has been shown to enhance stability against adversarial attacks \cite{bhojanapalli2021adversarial} and accuracy in medical data \cite{wu2021vision}.

\subsubsection{Robustness}
Recent studies show that Vision Transformers (ViTs) exhibit higher durability to attacks and disturbances compared to traditional CNNs, this is attributed to the self-attention mechanism, which allows the efficient exploitation of global image features \cite{bhojanapalli2021adversarial, paul2023are}. The study by Benz et al. \cite{bhojanapalli2021adversarial} found that ViTs require larger distortions to be fooled by white-box attacks (such as PGD and FGSM), while CNNs show vulnerabilities even at low levels of distortion. It was also found that adversarial examples generated in CNNs do not easily transfer to ViTs. The same study showed that Transformers rely more on low-frequency features of the image, which are tolerant to attacks, while CNNs showed that they depend more on high-frequency features, which are more vulnerable. At the same time, according to Bai et al. \cite{paul2023are} in a fair comparison between ViTs (e.g. DeiT-S \cite{touvron2021training}) and CNNs (e.g. ResNet-50 \cite{he2016deep}) of the same size and performance, ViTs show better resilience to adversarial attacks and generalize better to out-of-distribution data (such as the ImageNet-A \cite{hendrycks2021imagenet} and ImageNet-C \cite{hendrycks2019benchmarking} datasets). However, it was observed that the advantage of ViTs in robustness decreases when CNNs are trained with similar techniques to them. Finally, comparisons that included hybrid models that combine CNN and ViT features, namely local and global features, showed even higher robustness \cite{dey2023exploring}. 

\subsubsection{Medical Imaging}

Unlike natural image recognition and object detection, medical imaging presents unique challenges. Datasets are often much smaller and imbalanced, while disease patterns may be spread across different regions of the image rather than localized in well-defined objects \cite{shen2021overview}. This makes the global context modeling of Vision Transformers particularly useful, since self-attention can capture long-range dependencies that CNNs often miss. Based on a recent study, a key finding is that ViTs tend to be less affected by hidden stratification, that is, situations where the model bases its prediction on random or irrelevant features of the image instead of essential medical indicators \cite{chu2022visual}. This is of particular importance to diagnostic accuracy, as it reduces the risk of incorrect predictions due to bias in the data. Furthermore, in the study \cite{wu2021vision}, ViTs outperformed established CNNs in both accuracy and generalization for the classification of emphysema types from CT scans, demonstrating an improved ability to adapt to new and independent datasets. Similarly, in breast ultrasound image classification, ViT-B/32 achieved an accuracy of 86.7\% and an AUC of 0.95, surpassing leading CNNs in some cases, while also enabling better interpretability through the use of attention maps \cite{gheflati2022vision}.

\section{Evaluation of Object Recognition and Detection}
\label{Eval Object Recognition, Detection}

\subsection{Configuration}
The models we used for object recognition are shown in Table \ref{tab:classification_models}, while those we used for object detection are shown in Table \ref{tab:object_detection_models}. For each model in the corresponding process, we used the same framework, dataset, evaluation hyperparameters and GPU. 
\subsubsection{Pre-training}
For object recognition, most models are pre-trained on the ImageNet-1k dataset, except for the ViT models, which are pre-trained on the larger ImageNet-21k and then fine-tuned on ImageNet-1k, as they depend on pre-training on very large datasets followed by fine-tuning, as mentioned in \cite{dosovitskiy2021image}. In the case of object detection, all models are pre-trained on the Coco dataset, except for the YOLOS models, which are pre-trained on ImageNet-1k and then fine-tuned on Coco, as was originally proposed in \cite{fang2021yolos}. For evaluation we used the corresponding datasets' subsets for the two tasks. Experiments were carried out using the HuggingFace API and Google Colab, offering a Nvidia Tesla T4 GPU with 16GB VRAM. 

\subsubsection{Datasets}
For the Object Recognition task we used the ImageNet-1k dataset. ImageNet-1k \cite{imagenet-wiki}  is the most well-known subset of the ImageNet-21k and it’s used more widely. It includes 1,281,167 training images, 50,000 evaluation, and 100,000 test images, divided into 1,000 classes. For the purposes of this work, we only use the ImageNet-1k validation subset. 

For the Object Detection task we used the Coco dataset. Coco \cite{lin2014microsoft} is the most well-known benchmark for evaluating models in object detection. It is a large-scale publicly available dataset that includes images with detailed annotations for object detection and provides over 328,000 images of everyday scenes divided into 80 object categories. For the evaluation we only used the validation subset of Coco.

\subsection{Metrics}
For classification, model evaluation is done by measuring accuracy. Accuracy is the most common method for evaluating classification models. It is the ratio of the total number of correct predictions made by the model to the total number of predictions it made.

\[
\text{Accuracy} = \frac{\text{(Number of correct predictions)}}{\text{(Total number of predictions)}}
\]

For the evaluation in object detection, the metrics Precision, Recall, IoU and mAP are used.

\textbf{Precision:} Precision indicates the model's ability to make accurate positive predictions.

\[
Precision = \frac{TP}{TP + FP}
\]

\textbf{Recall:} Recall focuses on the model's ability to correctly identify positive samples from the entire set of positive cases.

\[
Recall = \frac{TP}{TP + FN}
\]

\textbf{Intersection over Union (IoU):} IoU is a metric that evaluates the extent of overlap between two bounding boxes, providing a measure of how well a predicted object aligns with its true counterpart. 

\textbf{Mean Average Precision (mAP):} In object detection, mAP measures the accuracy of placing bounding boxes by comparing them to the actual ones through the IoU metric. To calculate mAP, we first calculate the Average Precision (AP) for each category, creating precision-recall curves and finding the area under the curve (AUC). The final mAP score is the average of the AP scores for all categories, providing an overall performance index.

\[
mAP = \frac{1}{C} \sum_{i=1}^{C} AP_i
\]

\textbf{AP50} / \textbf{AP75}: They correspond to the average precision for specific IoU thresholds, for IoU=0.50 and IoU=0.75.

\textbf{APs, APm, APl:} AP metrics for small (APs), medium (APm), and large (APl) objects, based on the object’s size in the COCO dataset. They provide insight into how well a model performs on objects of different scales. The same applies to ARs, ARm, ARl.

\textbf{Average Recall (AR):} AR is the maximum recall given a fixed number of detections per image, averaged over categories and IoUs. AR1, AR10, AR100: The average recall given 1, 10, or 100 detections per image, respectively. These metrics measure the model’s ability to correctly find positive instances under varying maximum numbers of predictions.

\begin{table}[htbp]
\caption{CNN, Transformer, and Hybrid Models For Object Recognition Evaluation}
\begin{center}
\begin{tabular}{|l|c|c|c|}
\hline
Model & \#Params (M) & Image size & FLOPs (G) \\
\hline
\multicolumn{4}{|c|} {CNN} \\
\hline
ResNet-101 \cite{he2016deep} & 44.5 & $224^2$ & 7.9 \\
ResNet-152 \cite{he2016deep} & 60.2 & $224^2$ & 11.6 \\
EfficientNet-B0 \cite{tan2019efficientnet} & 5.3 & $224^2$ & 0.4 \\
EfficientNet-B4 \cite{tan2019efficientnet} & 19.0 & $380^2$ & 4.2 \\
EfficientNet-B7 \cite{tan2019efficientnet} & 66.3 & $600^2$ & 37.1 \\
\hline
\multicolumn{4}{|c|}{Transformer} \\
\hline
ViT-B/16 \cite{dosovitskiy2021image} & 86.6 & $224^2$ & 17.6 \\
ViT-L/16 \cite{dosovitskiy2021image} & 304.3 & $224^2$ & 61.6 \\
PvT-medium \cite{wang2021pyramid} & 44.2 & $224^2$ & 6.7 \\
PvT-large \cite{wang2021pyramid} & 61.4 & $224^2$ & 9.8 \\
Swin-B \cite{liu2021swin} & 87.8 & $224^2$ & 15.4 \\
Swin-L \cite{liu2021swin} & 196.5 & $224^2$ & 34.5 \\
\hline
\multicolumn{4}{|c|}{Hybrid} \\
\hline
CvT-21 \cite{wu2021cvt} & 31.6 & $224^2$ & 6.6 \\
CvT-21-384 \cite{wu2021cvt} & 31.6 & $384^2$ & 19.5 \\
LeViT-256 \cite{graham2021levit} & 18.9 & $224^2$ & 0.6 \\
LeViT-384 \cite{graham2021levit} & 39.1 & $384^2$ & 2.1 \\
ViT-Hybrid-Base \cite{dosovitskiy2021image} & 99 & $384^2$ & 49.6 \\
\hline
\end{tabular}
\label{tab:classification_models}
\end{center}
\end{table}

\begin{table}[htbp]
\caption{CNN, Transformer, and Hybrid Models For Object Detection Evaluation}
\begin{center}
\begin{tabular}{|l|c|c|c|}
\hline
Model & \#Params (M) & Image size & FLOPs (G)\\
\hline
\multicolumn{4}{|c|}{CNN} \\
\hline
Faster R-CNN \cite{ren2015faster} & 41.5 & $800^2$ & 134.7 \\
RetinaNet \cite{lin2017focal} & 33.8 & $800^2$ & 151.9 \\
SSD300 \cite{liu2016ssd} & 35.6 & $300^2$ & 35 \\
\hline
\multicolumn{4}{|c|}{Transformer} \\
\hline
Yolos-Tiny \cite{fang2021yolos} & 6.5 & $512 \times 768$ & 21.4 \\
Yolos-Base \cite{fang2021yolos} & 127.8 & $512 \times 768$ & 190.1 \\
\hline
\multicolumn{4}{|c|}{Hybrid} \\
\hline
Detr-ResNet-50 \cite{carion2020end} & 41.3 & $873 \times 1201$ & 102 \\
Detr-ResNet-101 \cite{carion2020end} & 60.2 & $873 \times 1201$ & 181.4 \\
Conditional-Detr-R50 \cite{meng2021conditional} & 43.2 & $873 \times 1201$ & 106.2 \\
Deformable-Detr \cite{zhu2021deformable} & 40 & $512 \times 768$ & 173 \\
\hline
\end{tabular}
\label{tab:object_detection_models}
\end{center}
\end{table}

\subsection{Object Recognition}
Table \ref{tab:classification_results} presents our evaluation results of various model architectures on the ImageNet dataset. As observed, the original Vision Transformer (ViT) achieves highly competitive performance compared to traditional CNNs, with ViT-Large reaching an accuracy of 82.5\%, surpassing landmark models such as ResNet and EfficientNet.However, this performance comes at the cost of a significantly increased number of parameters (304.3M) and high computational cost (61.6 GFlops).

Significant superiority is observed in the Swin Transformer models, where Swin-Large records the highest accuracy (86\%), surpassing all other models, both CNN and other ViT or hybrid models.This superiority is attributed to the Swin architecture, which combines the hierarchical feature extraction of CNNs with the shifted windows mechanism in self-attention, enabling the simultaneous exploitation of both global and local correlations within an image. Furthermore, Swin-Large maintains a significantly lower number of parameters (196.5M) and computational cost (34.5 GFlops) compared to ViT-Large.

The CvT (Convolutional Vision Transformer) models are also distinguished for their balance between performance and computational cost. For example, CvT-21-384 achieves an accuracy of 82.2\% with only 31.6M parameters and 19.2 GFlops, values much lower than those of the corresponding ViT and Swin models. The CvT architecture is based on the combination of convolutional and attention mechanisms, drawing advantages from both approaches and making the hybrid strategy particularly effective.

\begin{table}[htbp]
\caption{Object Recognition results sorted by increasing Accuracy number}
\centering
\scriptsize
\renewcommand{\arraystretch}{1.2} 
\setlength{\tabcolsep}{8pt}       
\begin{tabular}{|l|c|c|c|}
\toprule
Model & Acc. & \#Params (M) & FLOPs (G) \\
\midrule
Efficientnet-b0 & 0.72578 & 5.3 & 0.4 \\
Levit-256 & 0.80894 & 18.4 & 1.1  \\
CvT-21 & 0.81266 & 31.6 & 6.6  \\
PvT-Medium-224 & 0.81268 & 44.2 & 6.7  \\
ViT-Base-patch16-224 & 0.81675 & 86.6 & 17.6  \\
Pvt-Large-224 & 0.81560 & 61.4 & 9.9  \\
Efficientnet-b4 & 0.81612 & 19.0 & 4.2  \\
Resnet-101 & 0.81820 & 44.5 & 7.9  \\
Levit-384 & 0.82034 & 39.1 & 2.1  \\
Cvt-21-384 & 0.82294 & 31.6 & 19.5  \\
Vit-Large-patch16-224 & 0.82512 & \textbf{304.3} & \textbf{61.6}  \\
Resnet-152 & 0.82616 & 60.2 & 11.6  \\
Efficientnet-b7 & 0.83288 & 66.3 & 37.1  \\
Swin-Base-patch4-window7-224 & 0.84820 & 87.8 & 15.5  \\
ViT-hybrid-base-bit-384 & 0.84938 & 99.0 & 49.6  \\
Swin-Large-patch4-window7-224 & \textbf{0.86018} & 196.5 & 34.5  \\
\bottomrule
\end{tabular}
\label{tab:classification_results}
\end{table}

\subsection{Object Detection}

\begin{figure*}[htbp]
    \centering
    \begin{subfigure}{0.49\textwidth}
        \centering
        \includegraphics[width=\linewidth]{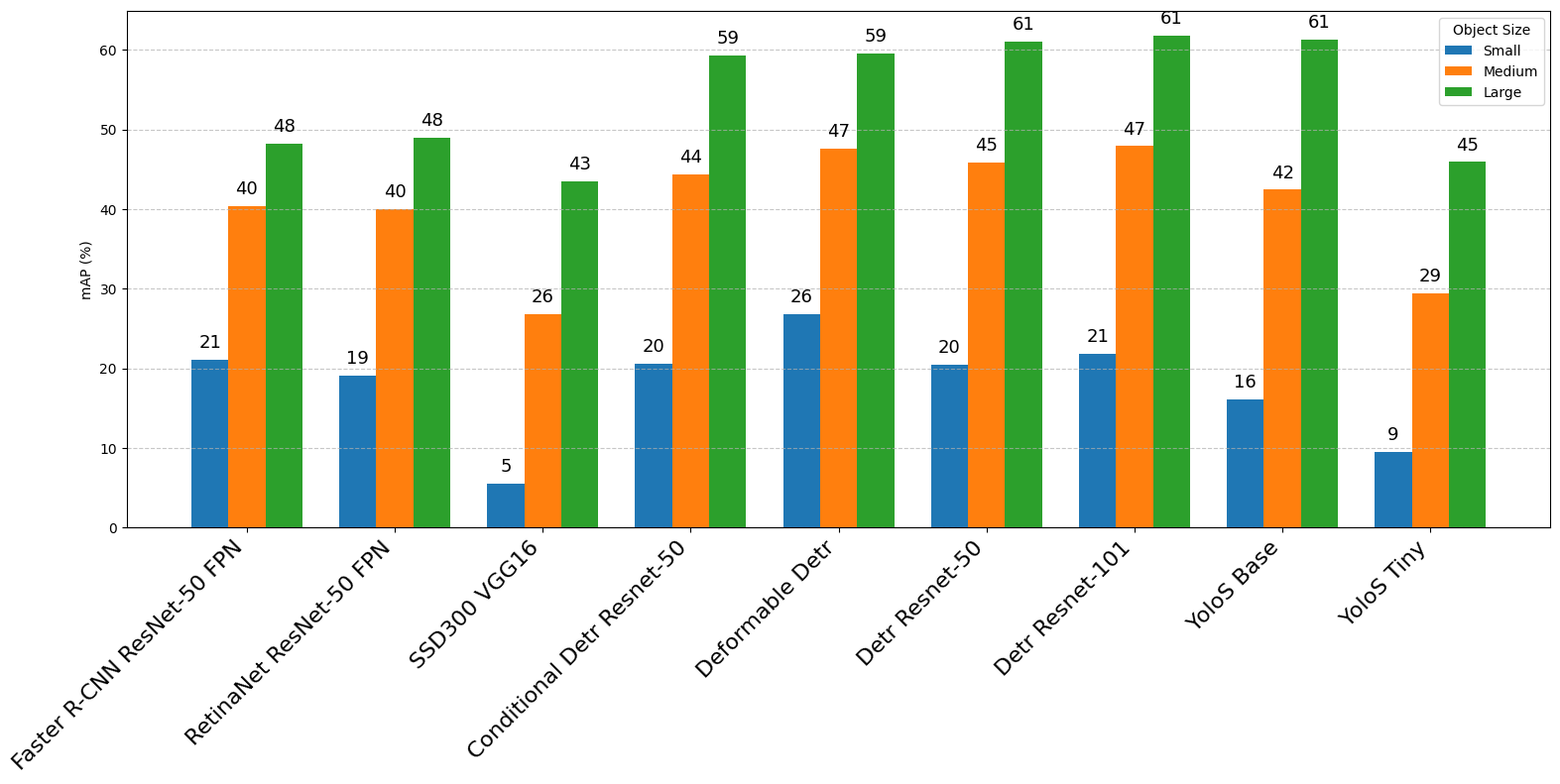}
        \caption{mAP per Object Size}
        \label{map per object size}
    \end{subfigure}
    \hfill
    \begin{subfigure}{0.49\textwidth}
        \centering
        \includegraphics[width=\linewidth]{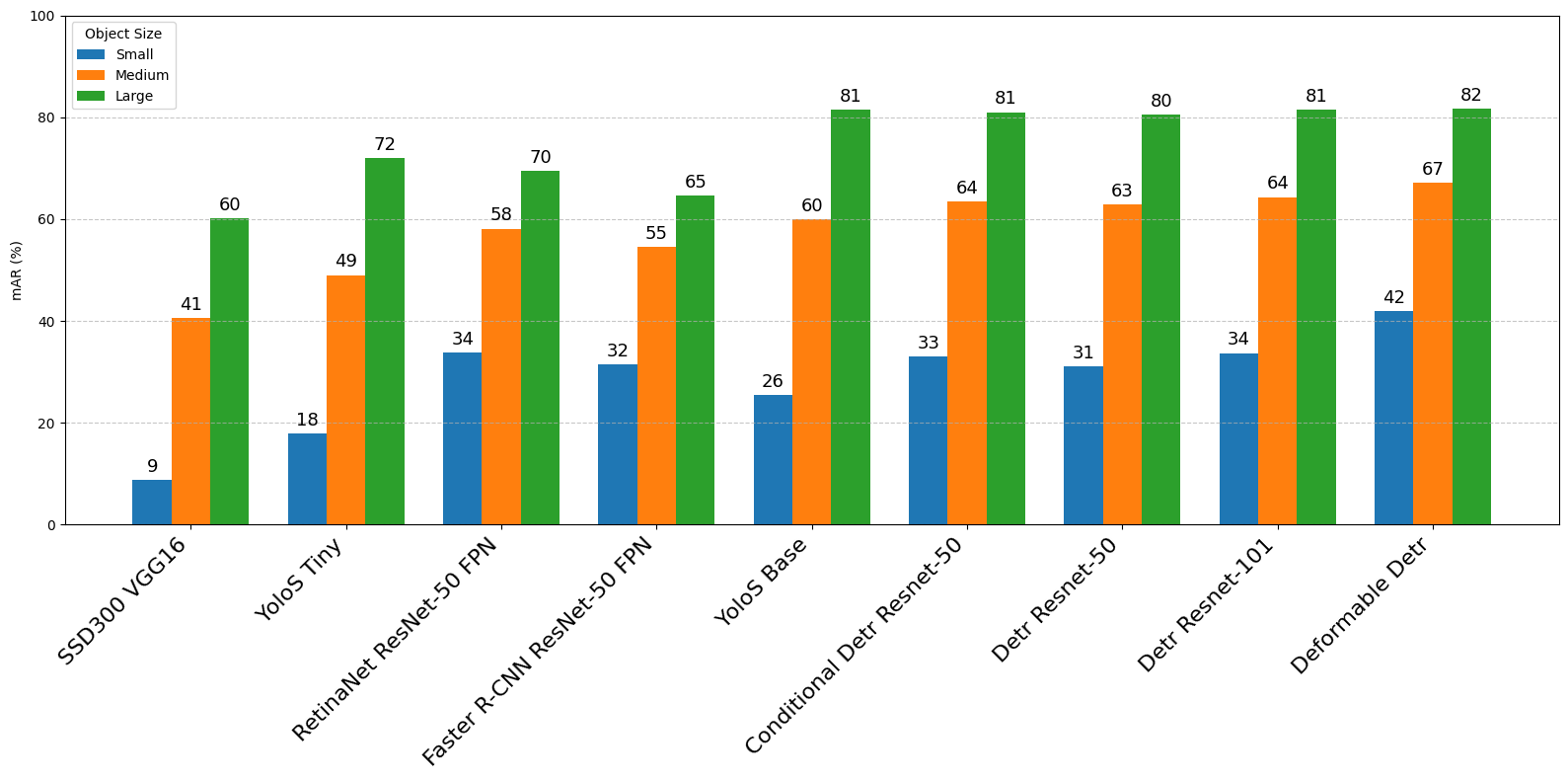}
        \caption{mAR per Object Size}
        \label{mAR per Object SIze}
    \end{subfigure}
    \caption{Results on COCO dataset: (a) mAP and (b) mAR per object size.}
    \label{fig:results}
\end{figure*}

Tables \ref{tab:Object_Detection_results} and \ref{tab:object_detection_ar} along with Fig. \ref{map per object size} and \ref{mAR per Object SIze} present the object detection results on the COCO dataset. Firstly, YOLOS—a pure Vision Transformer for object detection—manages to surpass several traditional CNN models in both mAP (39.4\%) and mAR (54.6\%), demonstrating the potential of transformer-based approaches in object detection. This significant result becomes even more noteworthy considering that YOLOS does not incorporate hierarchical or pyramidal feature processing (as CNNs do), but instead relies solely on a self-attention mechanism without inductive bias. This allows the model to better capture global dependencies in the image, resulting in high performance in the detection of large objects (mAPl 61.2\%, mARl 81.4\%). However, the lack of local processing significantly limits its ability to detect small objects (mAPs 16\%, mARs 26\%) as can be observed from Fig.\ref{map per object size}, \ref{mAR per Object SIze}.  Additionally, YOLOS shows increased parameter requirements (127.8M) and computational cost (190.1 GFlops), mainly due to the need for pre-training and fine-tuning, without necessarily justifying its superiority over other, more efficient models.

The DETR and Deformable DETR models, which adopt hybrid architectures, achieve even higher performance. DETR with a ResNet-101 backbone reaches an mAP of 43.4\% and an mAR of 59.0\%, while demonstrating leading results in the detection of large objects (mAPl 61\%, mARl 81\%), surpassing both established CNNs and YOLOS. This hybrid approach leverages both the global attention capabilities of transformers and the local information captured by CNNs, thereby enhancing the overall image understanding ability. Nevertheless, DETR still lags behind in the detection of small objects as demonstrated in Fig.\ref{map per object size}, \ref{mAR per Object SIze} (mAPs 21\%, mARs 34\%).

Deformable DETR stands out as the most efficient model, achieving the highest values in almost all performance metrics: mAP44.5\%, mAPs 26\%, mAPl 59\%, and mAR 62.9\%, mARs 40\%, mAPl 82\%. Its success is attributed to the deformable attention mechanism, which enables the model to focus on selected regions and across multiple scales, thereby enhancing the detection of small objects without compromising performance on medium and large objects.

\begin{table}[htbp]
\caption{Object Detection results sorted by increasing AP}
\centering
\scriptsize
\renewcommand{\arraystretch}{1.2} 
\setlength{\tabcolsep}{4pt}
\begin{tabular}{|l|c c c|cc|}
\toprule
Model & mAP & AP@50 & AP@75 & \#Params (M) & FLOPs (G) \\
\midrule
SSD300 & 0.251 & 0.416 & 0.263 & 35.6 & 35 \\
Yolos-Tiny & 0.287 & 0.472 & 0.289 & 6.5 & 21.4 \\
RetinaNet & 0.363 & 0.556 & 0.382 & 33.8 & 151.9 \\
Faster R-CNN & 0.369 & 0.585 & 0.398 & 41.5 & 134.7 \\
Yolos-Base & 0.394 & 0.592 & 0.414 & \textbf{127.8} & \textbf{190.1} \\
Conditional-Detr-ResNet-50 & 0.409 & 0.617 & 0.435 & 43.2 & 106.2 \\
Detr-ResNet-50 & 0.420 & 0.623 & 0.442 & 41.3 & 102 \\
Detr-ResNet-101 & 0.434 & \textbf{0.638} & 0.462 & 60.2 & 181.4 \\
Deformable-Detr & \textbf{0.445} & 0.636 & \textbf{0.486} & 40 & 173 \\
\bottomrule
\end{tabular}
\label{tab:Object_Detection_results}
\end{table}

\begin{table}[htbp]
\caption{Object detection results sorted by increasing AR}
\centering
\scriptsize
\renewcommand{\arraystretch}{1.2} 
\setlength{\tabcolsep}{4pt}
\begin{tabular}{|l|c c c|cc|}
\toprule
Model & mAR1 & mAR10 & mAR & \#Params (M) & FLOPs (G) \\
\midrule
SSD300 & 0.239 & 0.344 & 0.365 & 35.6 & 35 \\
Yolos-Tiny & 0.264 & 0.423 & 0.460 & 6.5 & 21.4 \\
RetinaNet & 0.313 & 0.499 & 0.539 & 33.8 & 151.9 \\
Faster R-CNN & 0.307 & 0.485 & 0.508 & 41.5 & 134.7 \\
Yolos-Base & 0.324 & 0.511 & 0.546 & \textbf{127.8} & \textbf{190.1} \\
Conditional-DETR-ResNet-50 & 0.334 & 0.540 & 0.581 & 43.2 & 106.2 \\
DETR-ResNet-50 & 0.333 & 0.532 & 0.574 & 41.3 & 102 \\
DETR-ResNet-101 & 0.344 & 0.548 & 0.590 & 60.2 & 181.4 \\
Deformable-DETR & \textbf{0.352} & \textbf{0.587} & \textbf{0.629} & 40 & 173 \\
\bottomrule
\end{tabular}
\label{tab:object_detection_ar}
\end{table}

\section{Evaluation on Medical Imaging}
\subsection{Configuration}
Best models per architecture group in table \ref{tab:classification_results} differ significantly in size. Therefore, for the experiments in this section we selected best representative models from each category that are close both in terms of performance as well as computational cost: Transformer (Swin-Base), Hybrid (CvT-21-384), and CNN (ResNet-152).

\subsubsection{Datasets}

For the Medical Imaging evaluation we used Chest-Xray14. The ChestX-ray14 \cite{wang2017chestx} is a large dataset of chest X-rays created to enhance deep learning applications in medical image analysis, and more specifically in the classification of common chest diseases. The dataset consists of 112,120 anterior chest X-rays extracted from 30,805 patients. It is divided into 14 chest diseases, selected based on frequency and diagnostic significance. Each image can also have multiple disease labels.

For the purposes of this study N-hot encoding was employed to convert text labels into arrays that represent the multiple labels each image may have. An N-hot encoded array represents the labels as a list of binary digits: “1” if the label applies to the image and “0” if it does not.

\subsubsection{Metrics}

For the Chest-Xray dataset we used the AUC-ROC metric.

AUC-ROC: The ROC (Receiver Operating Characteristic) curve and the AUC (Area Under the Curve) are important evaluation tools for the performance of classification models. The ROC curve is a graphical representation of the True Positive Rate (TPR) versus the False Positive Rate (FPR) for different decision thresholds. The AUC, or the area under the ROC curve, indicates how well a model can separate the categories, with a value close to 1 indicating high separation ability, while a value close to 0.5 indicates that the model does not separate the categories better than random selection. We use ROC-AUC because it is widely used to measure performance in multi-label classification tasks.

\subsubsection{Fine-tuning}
Fine-tuning was conducted for 3 epochs with a batch size of 16 for both training and evaluation, which provides a good balance between GPU memory usage and model convergence. Data loading was managed using 2 worker threads for increased efficiency, with the data loader set to drop the last batch if the dataset length was not divisible by the batch size. 

The optimizer utilized was AdamW, which is known for its effectiveness in fine-tuning transformer-based models. The learning rate was initialized at $2 \times 10^{-4}$, a value commonly adopted for such settings. Learning rate scheduling followed a cosine decay policy with a warmup ratio of 0.25; specifically, during the first 25\% of the training steps, the learning rate was gradually increased, which helps stabilize training before transitioning to the cosine decay phase. This strategy is widely used in transformer-based architectures to prevent divergence during the initial phase of training.

Logging was performed every 50 steps to monitor training progress, while model checkpoints were saved every 1,000 steps, enabling model recovery if necessary. Model selection was carried out by evaluating performance at the end of each epoch, and at the conclusion of training, the model with the highest ROC AUC score on the validation set was automatically selected and loaded. Additionally, the best-performing model was configured to be pushed to the HuggingFace Hub to ensure reproducibility.

\subsection{Optimization}

As shown in \cite{steiner2022how} ViTs exhibit limited performance when trained on small or insufficient datasets, such as those found in medical imaging as mentioned earlier in \cite{shen2017overview}. This limitation was attributed to their lack of inductive bias toward local image structures. The main technique proposed in \cite{steiner2022how} to address this issue is data augmentation. The data augmentation techniques we utilized are the following:

\subsubsection{Basic Random Augmentations}
\begin{itemize}
    \item Random Horizontal Flip: In each training epoch, every image has a 50\% chance of being flipped horizontally.
    \item Random Rotation: Each image is rotated by a random angle within a specified range [$-15^\circ$, $15^\circ$].
    \item Color Jitter: Each image undergoes a random change in brightness and contrast.
\end{itemize}

\subsubsection{CutMix}
CutMix is an augmentation technique that works as follows:

\begin{itemize}
    \item First, two images are randomly selected from the dataset.
    \item Then, a random rectangular patch of the same size and random position is cut from both images.
    \item The patch cut from the first image is placed in the empty area created in the second image, and vice versa for the first image.
    \item Finally, two new labels are created for the two new images, which are a linear combination of the two original labels.
\end{itemize}

\subsubsection{MixUp}
MixUp is an augmentation technique that creates a new sample by combining two existing ones and works as follows. Here, $x$ represents the data and $y$ the corresponding labels. The new sample $(\hat{x}, \hat{y})$ is constructed as:

\begin{align*}
    \hat{x} &= \lambda x_i + (1 - \lambda) x_j \\
    \hat{y} &= \lambda y_i + (1 - \lambda) y_j
\end{align*}

where $\lambda \in [0, 1]$ is the mixing coefficient. This means that the new image is intermediate between the two originals, and its label reflects the contribution of each original image.
   
Techniques such as CutMix and MixUp have demonstrated their effectiveness on ViT models in \cite{zhao2023mediaug} for medical imaging tasks. For this experiment, we utilized the Swin-Base model to evaluate these techniques on a variant of the ViT architecture, the model was trained for 10 epochs using the same hyperparameters as previously described. Furthermore, the technique of \textit{Early Stopping} was applied, terminating the training process if the ROC AUC score did not improve for three consecutive epochs.

\begin{table}[htbp]
\caption{Comparison of models on ChestX-ray14}
\centering
\scriptsize
\renewcommand{\arraystretch}{1.2} 
\setlength{\tabcolsep}{2pt}
\begin{tabular}{|l|c|c|c|c|c|}
\toprule
Model & \#Epoch & \#Params (M) & FLOPs (G) & Eval Loss & Eval ROC-AUC  \\
\midrule
Swin-Base            & 3 & \textbf{86.8} & 15.5 & 0.1676 & 0.8174  \\
Cvt-21-384-22k       & 3 & 31.2          & \textbf{19.2} & 0.1671 & \textbf{0.8219} \\
ResNet-152           & 3 & 58.2          & 11.6 & 0.1729 & 0.8113  \\
\bottomrule
\end{tabular}
\label{tab:model_comparison_Chest_Xray}
\end{table}

\begin{table}[htbp]
\caption{Comparison of Swin-Base data augmentatio techniques on ChestX-ray14}
\centering
\scriptsize
\renewcommand{\arraystretch}{1.2} 
\setlength{\tabcolsep}{6pt}
\begin{tabular}{|l|c|c|c|c|c|}
\toprule
Model Variant & \#Epoch & Eval Loss & Eval ROC-AUC \\
\midrule
Swin-Base                           & 3  & 0.1697 & 0.8174  \\
Swin-Base + Data Aug                & 10  & 0.1693 & 0.8430  \\
Swin-Base + CutMix                  & 9   & 0.1737 & 0.8290  \\
Swin-Base + MixUp                   & 10  & 0.1713 & 0.8361  \\
Swin-Base + Data Aug + MixUp        & 10  & \textbf{0.1690} & \textbf{0.8525}  \\
\bottomrule
\end{tabular}
\label{tab:swin_data_aug_comparison}
\end{table}

\subsection{Results and discussion}
From the results on the ChestX-ray14 dataset (Table \ref{tab:model_comparison_Chest_Xray}), it is observed that Transformer-based models outperformed the CNN, with the CvT-21-384 hybrid model in particular achieving the highest performance, with a ROC-AUC of 82.1\%, compared to 81.7\% for Swin-Base and 81.1\% for ResNet-152. This superiority of Transformer architectures aligns with recent findings \cite{gheflati2022vision, mancini2023visual, chu2022visual, wu2021vision}, which attribute their effectiveness to the modeling of global correlations in medical images through the self-attention mechanism.In such images, pathologies often manifest as diffuse or distant visual patterns that cannot be sufficiently captured by local filters alone, as used in CNNs \cite{wang2020deep}. CvT, by leveraging convolutional token embeddings in combination with self-attention, manages to combine the advantages of both CNNs and Transformers, achieving a better balance between local and global information. As a result, it performs better while having a significantly lower number of parameters and only slightly higher computational cost. This increased computational cost is due to CvT’s hybrid nature, performing both attention and convolution operations, and also because it processes higher resolution inputs (384×384) compared to the other two models (224×224). Swin also demonstrates competitive performance thanks to its hierarchical shifted window structure, which mimics the architecture of CNNs in order to also capture local dependencies.

From the evaluation of data augmentation techniques applied to the Swin-Base model on the ChestX-ray14 dataset (Table \ref{tab:swin_data_aug_comparison}), it becomes clear that augmentations significantly boost model performance. The combination of data augmentation techniques, specifically standard augmentation with MixUp, got the best result, by achieving a ROC-AUC of 85.25\%, outperforming the baseline Swin-Base model by approximately 4\%. Individual augmentation techniques, Data Augmentation alone (84.30\%) and MixUp alone (83.61\%), also show clear improvements, confirming the beneficial effect of each technique independently.However, CutMix demonstrates a lower performance (82.90\%), suggesting that its specific augmentation approach might be less effective for this particular dataset.

\section{Conclusions and future work}
It becomes evident that Vision Transformers (ViTs) can perform just as well or even better than traditional CNNs in tasks like image object recognition, object detection, and especially in medical imaging. The results showed that hybrid and hierarchical models like CvT and Swin Transformer manage to combine the best of both architectures, offering high accuracy with a good balance of computational cost. Also, by trying out data augmentation techniques such as the combination of Basic Augmentations and MixUp—especially on the Swin model for medical images—we saw a noticeable improvement in performance, proving that these methods can help ViTs on smaller datasets like ChestX-ray14. Overall, our study confirms that Transformers are very competitive in computer vision and medical tasks.

In the future, it would be worth exploring even more data augmentation techniques and investigate their impact on Transformer models, especially when applied to more challenging or imbalanced medical datasets. In addition, it is particularly important to evaluate more lightweight or optimized Transformer architectures, so that they can be effectively utilized in real-time or low-resource environments, such as on mobile devices or in hospitals with limited hardware capabilities.

\end{document}